\documentclass[10pt]{article} 
\usepackage[preprint]{tmlr}

\usepackage{amsmath,amsfonts,bm}

\def\eqref#1{equation~\ref{#1}}

\def\1{\bm{1}}

\DeclareMathAlphabet{\mathsfit}{\encodingdefault}{\sfdefault}{m}{sl}
\SetMathAlphabet{\mathsfit}{bold}{\encodingdefault}{\sfdefault}{bx}{n}

\usepackage{graphicx}
\usepackage[inline]{enumitem}
\usepackage[colorlinks=true,allcolors=blue
,citecolor=blue]{hyperref}
\usepackage{url}
\setcitestyle{numbers}

\title{Towards a Research Community in Interpretable Reinforcement Learning: the \texttt{InterpPol} Workshop}

\author{\name Hector Kohler \email \href{ mailto:hector.kohler@inria.fr}{hector.kohler@inria.fr} \\
      \addr CRIStAL, Université de Lille, Inria, CNRS, Centrale Lille, France 
      \AND
      \name Quentin Delfosse \email \href{ mailto:quentin.delfosse@cs.tu-darmstadt.de}{quentin.delfosse@cs.tu-darmstadt.de} \\
      \addr AIML Lab, Computer Science Department, Darmstadt, Germany
      \AND
      \name Paul Festor \email \href{ mailto:paul.festor19@imperial.ac.uk}{paul.festor19@imperial.ac.uk}\\
      \addr UKRI Centre for Doctoral Training in AI for Healthcare, Imperial College London, UK
      \AND
      \name Philippe Preux \email \href{mailto:philippe.preux@inria.fr}{philippe.preux@inria.fr} \\
      \addr CRIStAL, Université de Lille, Inria, CNRS, Centrale Lille, France
      }

\def\month{MM}  
\def\year{YYYY} 
\def\openreview{\url{https://openreview.net/forum?id=XXXX}} 

\begin{document}

\maketitle

\begin{abstract}
Embracing the pursuit of intrinsically explainable reinforcement learning raises crucial questions: what distinguishes explainability from interpretability? Should explainable and interpretable agents be developed outside of domains where transparency is imperative? What advantages do interpretable policies offer over neural networks? How can we rigorously define and measure interpretability in policies, without user studies? What reinforcement learning paradigms,are the most suited to develop interpretable agents? Can Markov Decision Processes (MDPs) integrate interpretable state representations? In addition to motivate an Interpretable RL community centered around the aforementioned questions, we propose the first venue dedicated to Interpretable RL: the \texttt{InterpPol} Workshop\footnote{\url{https://sites.google.com/view/interppol-workshop/home}}.
\end{abstract}

\section{Why care about interpretable RL?}\label{sec1}
Deep reinforcement learning (RL) agents are prone to suffer from a variety of issues that hinder them from learning optimal or generalizable policies. Prominent examples are \textit{reward sparsity}~\citep{andrychowicz2017hindsight} and \textit{difficult credit assignment}~\citep{Raposo2021SyntheticRF, Wu2023ReadAR} or goal misalignment~\citep{Koch2021-misalignment}, but not at test time. Such misalignments can be difficult to identify~\citep{Langosco2022goal}. 
A recently popular approach for identifying such shortcut behaviours is via methods from the field of eXplainable AI (XAI), whose aim is to identify the reasons explaining decisions made using a certain model~\citep{lapuschkin2019unmasking, GuidottiMRTGP19,SchramowskiSTBH20, RasXGD22,RoyKR22,SaeedO23}. 
Although XAI methods have helped us better understand, detect and correct many misaligned behaviours, they suffer from issues such as a lack of faithfulness, i.e the explanation does not faithfully present the model underlying decision process~\citep{HookerEKK19,ChanKL22,DeYoungJRLXSW20}, and coarse, low-level explanation semantics. For example, importance-map explanations indicate the importance of an input element for a decision without indicating \textit{why} this element is important~\citep{Kambhampati2021SymbolsAA, rightconcepts,TesoASD23}. 
\citep{Langosco2022goal, delfosse2024interpretable} show that heatmaps can be particularly misleading when applied to decisions made by RL agents. They argue that, learning rules-based policies defined over extracted intermediate object-centric and relational representations of states, results in intrinsically explainable agents: this allows for detecting and for correcting the previously mentioned problems. Hence one needs to go \textit{beyond explainability} and aim for interpretability (intrinsic explainability).

There have been mainly two approaches to learn interpretable policies~\citep{Milani2023ExplainableRL}. First, imitating neural network policies a posteriori with policies~\citep{luo2024insight} such as decision trees or programs~\citep{graf2024three, Delfosse2023InterpretableAE, guo2024efficient}, which are interpretable, i.e. explainable in nature~\citep{glanois2022survey}, allows for verifiability but often leads to complex policies (trees having a lot of nodes, hence hard to understand for a human being) as trees have to grow deep to represent neural policies~\citep{viper,verma2018programmatically}. 
These approaches rely on object extraction method such as FastSAM~\citep{zhao2023fast} or SPACE+MOC~\citep{delfosse2021moc}.
Second, recent approaches have directly optimized interpretable policies with reinforcement learning leading to less complex solutions than imitated policies~\citep{pmlr-v108-silva20a,custard} but suffer from stability issues, lack of theoretical grounding, or require inductive bias~\citep{kohler2024limits}. Furthermore, independent of the approach to learn interpretable policies, these policies need to be defined over semantically meaningful state spaces, e.g not pixels but objects in images in the case of Atari games. This calls for, not only policy learning algorithms, but also for interpretable MDPs learning algorithms in the likes of AlignNet~\citep{creswell2020alignnet}.

The biggest challenge in interpretable RL research remains the lack of definitions and a common paradigm. In the general machine learning (ML) litterature, there have been attempts at quantifying how interpretable a model is either with metrics such as \textit{simulatability} (time necessary for a human to compute the output of a model)~\citep{interpret_Lipton16a}, or with user studies~\citep{chen2022use}. But there is no similar work specifically for RL tasks. There is also a lack of tools for comparing interpretable policies from different classes, e.g comparing a tree and a program.

Despite research on interpretable RL being at an early stage, there have been encouraging applications of the latter in fields like healthcare~\citep{festor2021enabling, finalePreClinical}. Research is also carried to understand the link between XRL and the influence it has on the final decision maker~\citep{nagendran2023quantifying}. Interpretability has to be developed with a user-centric approach and raises questions of trust between users and AI. Benchmarks to evaluate interpretable methods are e.g \texttt{GymDSSAT}~\citep{gautron2022gymdssat} and OCAtari~\citep{delfosse2023ocatari}.

\section{Venues for Interpretable RL}
\subsection{Existing venues}
\begin{figure}[t]
    \centering
    \includegraphics[width=0.6\linewidth]{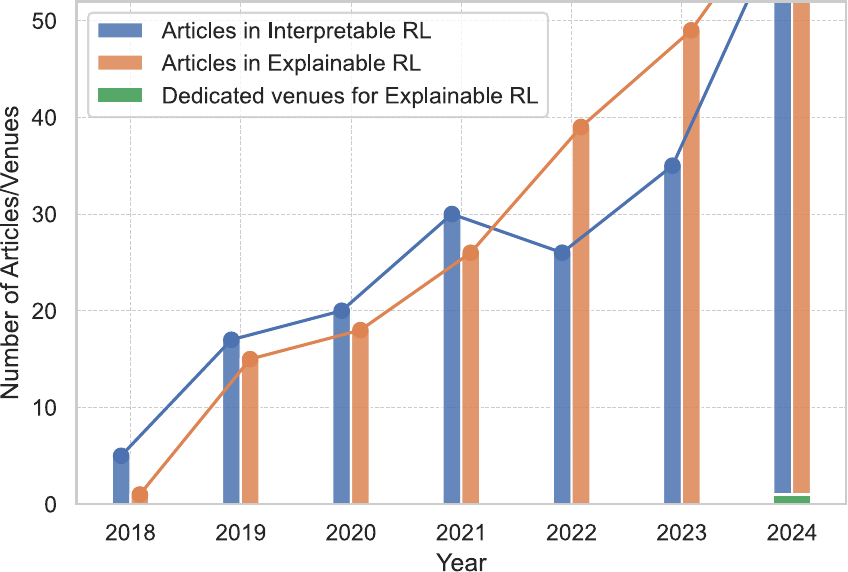}
    \caption{\textbf{We need XRL/IRL workshops.} The articles with "interpretable/explainable reinforcement learning" in the title against number of dedicated venues on XRL, using \texttt{google scholar}.}
    \label{fig:community}
\end{figure}

There is a need for crystallizing both a community and a set of research directions. As shown in Figure~\ref{fig:community}, there is a lack dedicated venues for people interested in explainable and interpretable RL: even though it is a growing field in terms of published papers, there has been only one dedicated venue \footnote{\url{https://xai4drl.github.io/}}.
\textbf{The community needs definitions and a formal research paradigm}. 

\subsection{\texttt{InterpPol} Workshop: an attempt to crystalize an Interpretable RL community}
Motivated by the need for research in Interpretable RL described in section \ref{sec1} and the growing publications on the topic shown in figure \ref{fig:community}, we organize the first dedicated workshop on Interpretable RL: the Workshop on Interpretable Policies in Reinforcement Learning, \texttt{InterpPol}\footnote{\url{https://sites.google.com/view/interppol-workshop/home}}. \texttt{InterPpol} will take place on August 9 at the Reinforcement Learning Conference in Amherst, MA \footnote{\url{https://rl-conference.cc/index.html}}. 


The purposes of the workshop on Interpretable Policies in RL (\texttt{InterpPol}) are to create a community as well as better formalize and crystallize the problem of learning interpretble policies. The core topics of interest at \texttt{InterpPol} are (but not limited to):

\begin{itemize}[leftmargin=8mm]
\item \textit{why learn interpretable policies?} What are the interpretability requirements of sequential decision problems, e.g in the health sector? Are there practical advantages of learning interpretable policies such as decision trees rather than neural networks with respect to RL goal alignment? Why looking for intrinsically explainable --interpretable-- policies rather than post-hoc explainability?

\item \textit{How to define interpretability?} What is an interpretable policy? How to compare different classes of interpretable policies such as trees or programs? How to quantify interpretability of policies with metrics or with (simulated) user studies?

\item \textit{How to learn interpretable policies?} What advantages does each learning paradigms (e.g imitation learning, direct reinforcement learning, or evolutionary methods) incorporate? When to favor learning interpretable policy rather than distilling? Do certain interpretable structures prove easier to learn than others, e.g trees against programs? How to link combinatorial optimization and interpretable RL?

\item \textit{What sequential problems can be solved with interpretable RL?} How can we refine MDPs to be most conducive to learning interpretable policies, especially in terms of interpretable state space representations? How to learn new models over interpretable state spaces? Do all interpretable MDPs have an interpretable solution? How to use reward shaping to trade-off between interpretability and performance of the learned policy?
\end{itemize}
\noindent
We also call for connex topics that could lead to collaborations such as real-world data RL algorithms and benchmarks if they can be adapted to be interpretable. \\

\noindent 
Please submit your papers on \url{https://openreview.net/group?id=rl-conference.cc/RLC/2024/Workshop/InterpPol} and do not hesitate to contact us if you can help reviewing. 
\textbf{The submission deadline will be 26th of April and decisions will be released on 20th of May. }
Authors from key papers of section \ref{sec1} will also talk during the workshop and be present at poster sessions as shown in the \texttt{InterpPol} schedule \footnote{\url{https://sites.google.com/view/interppol-workshop/program?authuser=0}}.

\subsection{Beyond the workshop}
We aim to create an open community on Interpretable RL. We hope to draft a position paper during \texttt{InterpPol} that advocate for advancements and standards in the interpretable and explainable field, such as the one done by ~\cite{sucholutsky2023getting} on representation alignment, or by~\cite{mannor2023towards} on deployment. 
We will establish an open Google Group on Interpretable RL \footnote{\url{https://groups.google.com/g/interpretablerl}} to exchange ideas, research, and developments in the field of interpretable reinforcement learning. This group aims to foster collaboration and discussion among researchers, practitioners, and enthusiasts who are keen on advancing our understanding and application of interpretability in reinforcement learning algorithms.
Finally, we hope to organize a series of online seminars on the topics of interest in the like of \url{https://sites.google.com/view/rltheoryseminars}.

If you are interested in Interpretable RL contact us at \url{mailto:interppol.workshop@gmail.com}

\bibliography{main}
\bibliographystyle{tmlr}


\end{document}